\documentclass{article} % For LaTeX2e
\usepackage{iclr2026_conference,times}

\usepackage{amsmath,amsfonts,bm}

\def\eqref#1{equation~\ref{#1}}
\def\1{\bm{1}}

\DeclareMathAlphabet{\mathsfit}{\encodingdefault}{\sfdefault}{m}{sl}
\SetMathAlphabet{\mathsfit}{bold}{\encodingdefault}{\sfdefault}{bx}{n}

\usepackage{hyperref}
\usepackage{url}
\usepackage{graphicx}
\usepackage{booktabs}
\usepackage{wrapfig}
\usepackage{subcaption}
\usepackage{pifont}
\usepackage{enumitem}

\title{Exploiting Tree Structure for Credit Assignment in RL Training of LLMs}

\author{
Hieu Tran\thanks{Equal contribution} $^{1,2}$, 
Zonghai Yao\footnotemark[1] $^{1,2}$, 
\textbf{Hong Yu$^{1,2,3}$}\\
$^{1}$Center for Healthcare Organization and Implementation Research, VA Bedford Health Care  \\
$^2$Manning College of Information and Computer Sciences, University of Massachusetts Amherst\\
$^3$Miner School of Computer and Information Sciences, University of Massachusetts Lowell\\
}

\iclrfinalcopy % Uncomment for camera-ready version, but NOT for submission.
\begin{document}

\maketitle

\begin{abstract}
Reinforcement learning improves LLM reasoning, yet sparse delayed reward over long sequences makes token-level credit assignment the key bottleneck. 
We study the verifiable-reward setting, where the final answer is checkable and multiple responses can be drawn per prompt. 
Reasoning tasks in math and medical QA align with this setup, where only a few decision tokens significantly impact the outcome. 
PPO offers token-level advantages with a learned value model, but it is complex to train both the actor and critic models simultaneously, and it is not easily generalizable, as the token-level values from the critic model can make training prone to overfitting.
GRPO is critic-free and supports verifiable rewards, but spreads a single sequence-level return across tokens and ignores branching. We introduce \textbf{Prefix-to-Tree (P2T)}, a simple procedure that converts a group of responses into a prefix tree and computes \emph{nonparametric} prefix values \(V(s)\) by aggregating descendant outcomes. 
Built on P2T, we propose \textbf{TEMPO} (\emph{\textbf{T}ree-\textbf{E}stimated \textbf{M}ean Prefix Value for \textbf{P}olicy \textbf{O}ptimization}), a critic-free algorithm that augments the group-relative outcome signal of GRPO with \emph{branch-gated} temporal-difference corrections derived from the tree. 
At non-branch tokens, the temporal-difference (TD) term is zero, so TEMPO reduces to GRPO; at branching tokens, it supplies precise token-level credit without a learned value network or extra judges/teachers. 
On Qwen3-1.7B/4B, TEMPO outperforms PPO and GRPO on in-distribution (MATH, MedQA) and out-of-distribution (GSM-HARD, AMC23, MedMCQA, MMLU-Medical) benchmarks, and reaches higher validation accuracy with roughly the same wall-clock time.
\footnote{Our code can be accessed at: \url{https://github.com/fatebreaker/tempo}}

\end{abstract}

\begin{quote}
\small
“In chess, a tempo is a move saved at a fork; in learning, credit should fall on that move.”
\end{quote}

\section{Introduction}

Reinforcement learning (RL)~\citep{sutton1998reinforcement} is an effective way to strengthen the reasoning of large language models (LLMs) \citep{zhang2025survey}. 
In LLM settings, rewards are sparse and delayed and sequences are long \citep{jaech2024openai,guo2025deepseek}, so the key challenge is \textbf{credit assignment}: give the outcome reward to the few tokens that really change the solution. 
We study the \emph{verifiable-reward} setting, where the final answer for a prompt is checkable and we can draw multiple responses for the same prompt. This is common in long “thinking” or chain-of-thought (CoT) tasks such as mathematics and medical QA, where most steps are low-impact and only a small set of \emph{decision tokens} (e.g., strategy choice, formula selection, diagnostic commitment) moves the outcome. 
A good learning rule should use the branching structure across responses and focus credit on those decision points.

\textbf{PPO} gives token-level advantages with a learned value and generalized advantage estimation (GAE), which mixes Monte Carlo (MC) returns with temporal-difference (TD) bootstrapping \citep{schulman2015high,schulman2017proximal}. However, jointly training the actor and critic is complex and often fails to generalize, as critic-derived token-level values can induce overfitting.
\textbf{GRPO} removes the critic and uses group-relative baselines over responses to the same prompt \citep{shao2024deepseekmath,yu2025dapo}. It is simple and fits verifiable rewards. However, it spreads a single sequence-level signal across all tokens and overlooks mid-trajectory decisions. As a result, token-level credit is weak when reasoning branches. 
Recent ``key-token ideas''~\citep{wang2025beyond} move toward finer signals, but they do not use the \textbf{implicit prefix tree} that multiple responses to the same prompt already define.

\begin{wrapfigure}{r}{0.4\textwidth}
    \centering
    \includegraphics[width=\linewidth]{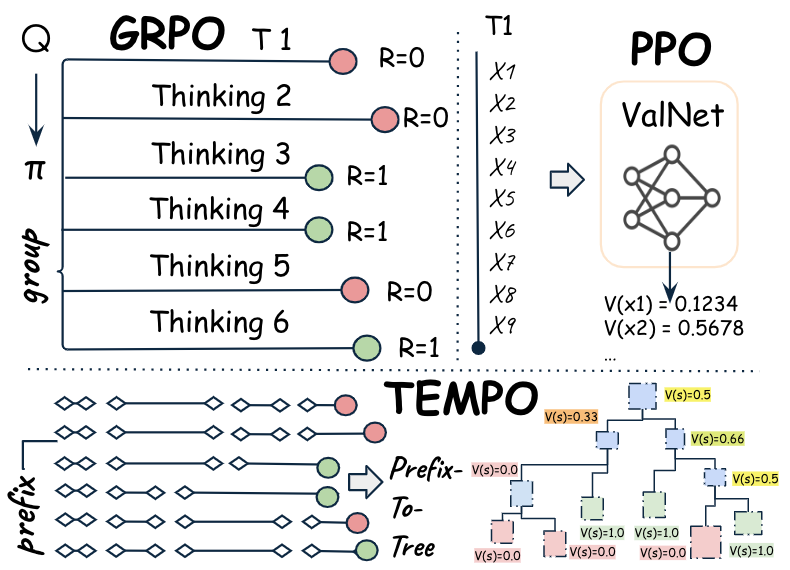}
    \caption{\textbf{Comparison of credit assignment for RL training with verifiable rewards.}
    GRPO: all tokens in each sampled answer share one sequence-level return; branching is ignored so credit spreads evenly.
    PPO: a learned value network estimates \(V(s_t)\) and provides token-level advantages via GAE, but requires a critic and higher compute.
    TEMPO: convert the answer group for one prompt into a prefix tree and compute \emph{nonparametric} prefix values \(V(s)\) by averaging descendant outcomes; use \emph{branch-gated} TD corrections to assign credit at branches.}
 \vspace{-4mm}
\label{fig:comparison}
\end{wrapfigure}

We present \textbf{Prefix-to-Tree (P2T)} as a simple procedure that converts a group of responses into a prefix tree and computes \emph{nonparametric} prefix values $V(s)$ by averaging descendant returns. 
Building on P2T, we introduce \textbf{TEMPO} (\emph{\textbf{T}ree-\textbf{E}stimated \textbf{M}ean Prefix Value for \textbf{P}olicy \textbf{O}ptimization}), a critic-free policy optimization method that restores token-level credit only where it matters. 
For each prompt, sampled responses form paths in the implicit prefix tree. 
TEMPO augments the group-relative outcome signal of GRPO with \emph{branch-gated} temporal-difference (TD) corrections derived from the tree: at non-branching tokens $V(s_{t+1}) = V(s_t)$ and the TD error is zero, so the update reduces to GRPO, while at branching tokens it supplies precise token-level credit. 
TEMPO maintains the GRPO training loop and cost. 
It does not train a value model or add a process reward model or a judge, it also does not require a teacher or a new sampler.

\paragraph{Empirical scope and applicability.}
Across \textit{Qwen3-1.7B} and \textit{Qwen3-4B}, TEMPO consistently attains higher accuracy than PPO~\citep{schulman2017proximal}, GRPO~\citep{shao2024deepseekmath}, and HEPO~\citep{wang2025beyond} on both in-distribution (MATH, MedQA) and out-of-distribution (GSM-HARD, AMC23, MedMCQA, MMLU-Medical) evaluations, while reaching strong validation performance in less wall-clock time. 
Validation curves indicate that, on math reasoning, approaches that emphasize token-level structure, such as HEPO~\citep{wang2025beyond} (e.g., focusing updates on high-entropy decision tokens) already enjoy an advantage, suggesting the RL phase mainly reinforces reasoning patterns learned during pretraining and SFT. 
Yet, TEMPO goes further by injecting tree-gated TD credit at the exact branching points. 
On medical reasoning, where domain knowledge must be newly acquired, methods that rely on group-relative exploration such as GRPO~\citep{shao2024deepseekmath} generalize better than purely exploitation-oriented updates; TEMPO combines this robust group baseline with branch-aware TD from P2T’s nonparametric prefix values, improving both convergence speed and final accuracy. 
In practice, TEMPO is most beneficial when rewards are verifiable and prompts yield meaningful branching, delivering precise token-level credit without a value network or auxiliary judges, and serving as a drop-in, efficiency-preserving upgrade to GRPO-style training. Our key \textbf{contributions} include:

\begin{enumerate}
[leftmargin=.1in,topsep=0.2pt]
    \item  We introduce \textbf{Prefix-to-Tree (P2T)}, a simple procedure that converts each prompt’s group of responses into a prefix tree and derives \emph{nonparametric} prefix values $V(s_t)$ by aggregating descendant outcomes.
    \item Building on P2T, we propose \textbf{TEMPO}, a drop-in, GRPO-compatible algorithm that augments the group-normalized outcome signal with \emph{branch-gated} TD corrections, providing precise token-level credit at decision points while retaining GRPO-like compute and simplicity.
    \item On \textit{Qwen3-1.7B/4B}, TEMPO improves convergence speed and final accuracy over other baselines on in-distribution (MATH, MedQA) and out-of-distribution (GSM-HARD, AMC23, MedMCQA, MMLU-Medical) benchmarks under the same hardware budget.
\end{enumerate}

\section{Related Work}
Credit assignment is central in post-training for reasoning LLMs. RLHF brought PPO with a learned value (critic) and GAE to reduce variance \citep{ouyang2022training,schulman2017proximal,schulman2015high}.
This improved alignment, however, comes at the cost of critic training, which adds complexity and tuning, and value prediction is brittle on long chains. 
To avoid a critic, several lines move towards value-free or RL-free updates that treat the entire response as a single action. 
DPO optimizes pairwise preferences in an offline bandit view \citep{rafailov2023direct}. 
Rejection-sampling methods, such as RestEM, fine-tune only on full high-reward responses \citep{singh2023beyond}.
RLOO, GRPO, and DAPO compute group-normalized sequence advantages over multiple samples of the same prompt, thereby removing the value network \citep{ahmadian2024back,shao2024deepseekmath,yu2025dapo}. 
These methods are simple and stable, but their feedback is sequence-level and credits all tokens equally, which weakens token-level credit in long reasoning.

A second thread tries to push feedback below the sequence. Token or span-level preference and dense-reward methods give finer signals \citep{yoon2024tlcr,yang2024selective,chan2024dense}.
Process supervision verifies intermediate steps or chain consistency to localize the first error \citep{lightman2023let,chen2024step,setlur2024rewarding,chen2024alphamath,zhang2024rest}.
Yet many step-by-step or tree-style approaches depend on a learned process reward model (PRM) or a judge to score nodes, which re-introduces a value function and adds verifier training cost. 
Some recent work also injects ad-hoc Monte Carlo (MC) signals into DPO to flag faulty steps \citep{hwang2024self,setlur2024rl}.
Our approach follows the value-free direction but uses the trajectory’s structure: it forms non-parametric prefix values from sibling continuations within a prompt group, then applies temporal-difference updates only at branching tokens where returns diverge, while non-branch tokens fall back to a GRPO-like baseline. 
In this way the update is simple like GRPO, yet it concentrates credit on decision points without a PRM.

Several contemporaneous works exploit the \emph{tree structure} of rollouts to densify credit assignment and/or cut sampling cost. TreePO \cite{li2025treepo} reframes on-policy rollouts as a tree search with segmented decoding and heuristic branching/fallback, amortizing shared prefixes (KV caching) and introducing a tree-based segment-level advantage estimator; this improves stability and reduces sampling compute while maintaining or improving accuracy. TreeRPO \cite{yang2025treerpo} extends GRPO by performing explicit tree sampling and forming step-level sibling groups to estimate expected rewards per step, yielding dense process signals and reporting consistent gains over GRPO with shorter responses. TreeRL \cite{hou2025treerl} integrates an entropy-guided sampler (EPTree) that branches at uncertain tokens, then back-propagates leaf rewards to provide global and local (step) advantages thereby eliminating a separate process reward model. Tree-OPO \cite{huang2025tree} leverages \emph{off-policy} teacher MCTS to build prefix trees and proposes staged, prefix-conditioned advantage estimation to stabilize GRPO-style updates. Unlike methods that require dedicated tree samplers (TreeRPO/TreeRL) or off-policy teacher trees (Tree-OPO), TEMPO operates in the standard GRPO setting and treats the \emph{implicit} prefix tree formed by a group of responses as a nonparametric value baseline: it computes $V(s_t)$ from all completions sharing the prefix $s_t$ and adds a token-level TD correction to the group-relative (Monte Carlo) signal. This yields branch-aware advantages without a learned value network, extra reward/process models, or special search procedures, while remaining fully on-policy and drop-in compatible with GRPO training loops.

\section{Preliminaries}

We begin by reviewing the advantage estimation used in Proximal Policy Optimization and Group Relative Policy Optimization.

\paragraph{PPO.} 
PPO \cite{schulman2017proximal} is a policy gradient method that stabilizes updates via a clipped objective. A key component is the estimation of the advantage function $A_t$, which measures how much better an action $a_t$ is compared to the average action at state $s_t$. PPO commonly employs \emph{generalized advantage estimation} (GAE) \cite{schulman2017proximal}, defined as
\[
\hat A_t^{\mathrm{GAE}(\gamma,\lambda)}
=\sum_{l=0}^{T-t-1} (\gamma\lambda)^l \,\delta_{t+l},
\qquad
\delta_t = r_t + \gamma V(s_{t+1}) - V(s_t).
\]

In the original formulation, $\gamma$ serves as a discount factor that reduces the weight of delayed rewards and helps stabilize infinite-horizon settings.  
However, in the context of large language model (LLM) training, it is common to set $\gamma=1.0$ so that long completions are not penalized relative to short ones.  
With this setting, the GAE formula simplifies to
\[
\hat A_t^{\mathrm{GAE}(\lambda)}
=\sum_{l=0}^{T-t-1} \lambda^l \,\delta_{t+l},
\qquad
\delta_t = r_t + V(s_{t+1}) - V(s_t).
\]

This version differs from the original in that there is no temporal discounting; all rewards contribute equally regardless of their position in the sequence. The bias–variance tradeoff is then controlled solely by the parameter $\lambda$.

\textbf{Special cases.}
\begin{itemize}
\item \emph{$\lambda=0$ (TD(0)).} 
\[
\hat A_t^{\lambda=0}=\delta_t
= r_t + V(s_{t+1}) - V(s_t),
\]
the one-step temporal-difference error (lowest variance, highest bias).

\item \emph{$\lambda=1$ (Monte Carlo).}
\[
\hat A_t^{\lambda=1}
=\sum_{l=0}^{T-t-1} r_{t+l} + V(s_T) - V(s_t).
\]
If $s_T$ is terminal so $V(s_T)=0$, then
\[
\hat A_t^{\lambda=1}
=\Big(\sum_{l=0}^{T-t-1} r_{t+l}\Big) - V(s_t),
\]
i.e., the full Monte Carlo return minus the baseline (unbiased, higher variance).
\end{itemize}

% This estimator interpolates between temporal-difference (TD) and Monte Carlo (MC) returns: $\lambda=0$ reduces to TD(0), while $\lambda=1$ recovers the MC return. In practice, intermediate values such as $\lambda=0.95$ balance bias and variance. Importantly, PPO requires training a separate value model $V(s)$ to compute $\delta_t$.

\paragraph{GRPO.}
GRPO \cite{shao2024deepseekmath} was designed for reinforcement learning with verifiable feedback. 
Formally, for each question $q$, a group of $G$ responses $\{o_1, \dots, o_G\}$ is sampled from the old policy $\pi_{\theta_{\text{old}}}$, and a reward model assigns scores $r=\{r_1,\dots,r_G\}$. 
These rewards are then normalized by subtracting the group mean and dividing by the group standard deviation. 
Under outcome supervision, the normalized reward $\tilde r_i = \frac{r_i - \text{mean}(r)}{\text{std}(r)}$ is applied uniformly to all tokens of output $o_i$, so that
\[
\hat A_{i,t} = \tilde r_i = \frac{r_i - \text{mean}(r)}{\text{std}(r)}, \quad \forall t \in o_i.
\]
Under process supervision, the same normalization is applied at the step level, and the normalized step rewards are distributed to the corresponding tokens. GRPO thus avoids training a separate value model and provides efficient group-relative baselines, but it relies purely on Monte Carlo outcomes and discards token-level temporal structure.
This setup highlights the gap: PPO provides token-level advantages but requires a value model, while GRPO is model-free but trajectory-level only. Our method, TEMPO, combines the strengths of both.

\begin{figure*}[t]
    \centering
    \includegraphics[width=\linewidth]{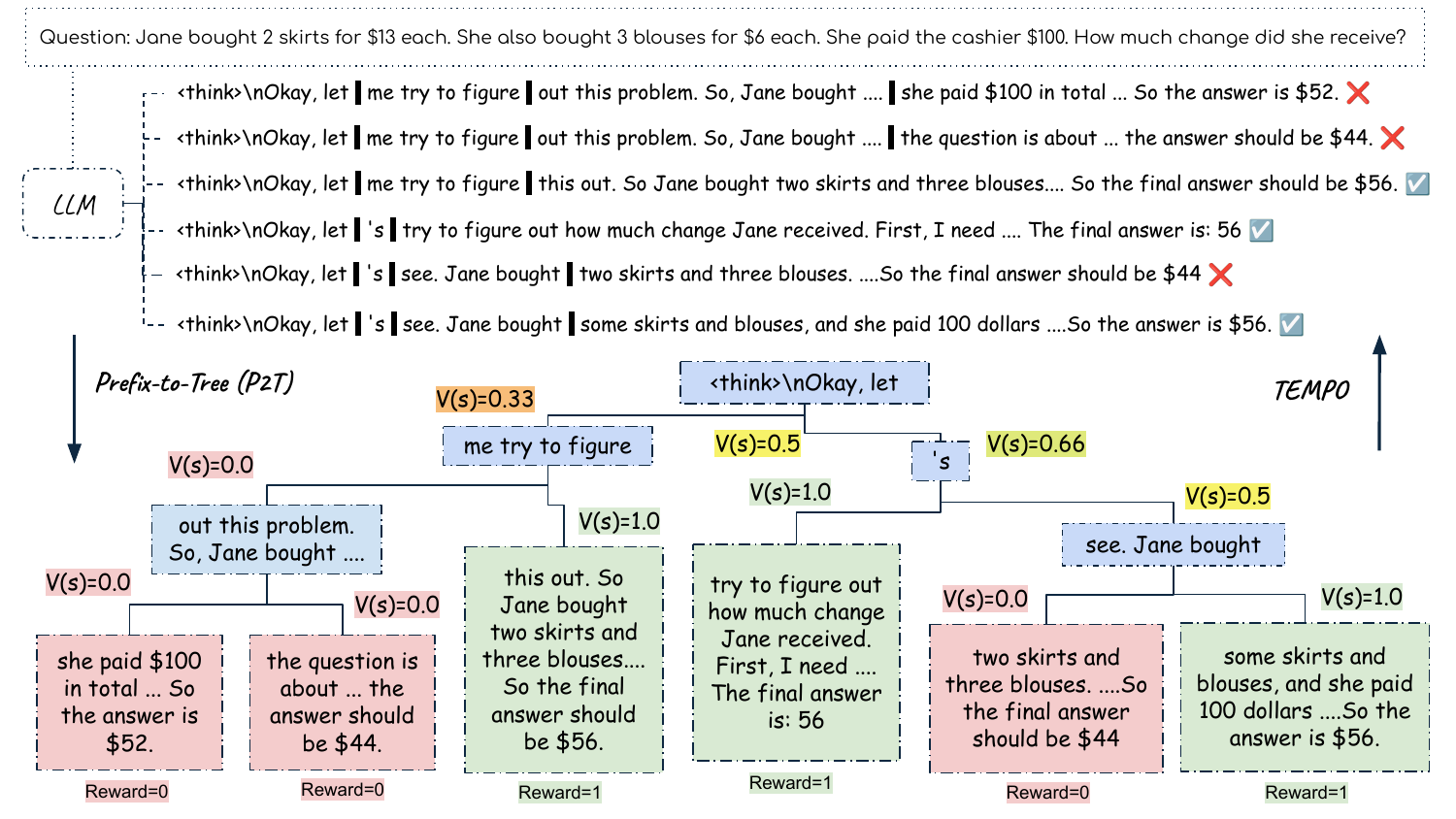}
    \caption{Overview of prefix tree value estimation in TEMPO. Each node corresponds to a token prefix $s$, with $V(s)$ estimated by averaging over the outcomes of all descendant completions. Green leaves denote correct responses ($r=1$), red leaves denote incorrect ones ($r=0$). Intermediate nodes inherit averaged values (e.g., $V(s)=0.5$), providing informative signals at branching points.}
        % \vspace{-5mm}
    \label{fig:overview}
\end{figure*}

\section{Methodology}

\subsection{Value Estimation from Prefix Tree}

Figure~\ref{fig:overview} illustrates how TEMPO derives value estimates directly from the tree structure formed by a group of sampled responses. Each path in the tree corresponds to a response generated by the policy, and each node represents a token prefix $s_t$ up to time $t$. The tree branches whenever different responses diverge at a given token. Terminal nodes are assigned rewards $r \in \{0,1\}$ based on verifiable correctness (e.g., whether the final answer matches the ground truth).

Instead of training a separate value model as in PPO, TEMPO computes $V(s_t)$ directly from the group of trajectories. For a given prefix $s_t$, the value is estimated as the average normalized reward of all descendant completions that share this prefix:
\[
V(s_t) \;=\; \frac{1}{|D(s_t)|} \sum_{j \in D(s_t)} r_j,
\]
where $D(s_t)$ is the set of responses passing through $s_t$, and  $r_j$ is the outcome reward. This provides a \emph{value function} without introducing an additional learned critic.

In the example shown in Figure~\ref{fig:overview}, some prefixes lead to correct answers ($r=1$) while others lead to incorrect ones ($r=0$). TEMPO propagates these signals upward by averaging over the subtree, yielding intermediate values (e.g., $V(s_t)=0.5$ when half of the descendant completions are correct). As a result, branch nodes obtain informative value estimates that reflect the quality of their continuations, while non-branch nodes naturally inherit consistent values from their unique continuation. This design ensures that tokens along successful reasoning paths (green leaves) contribute positively to the estimated value of their prefixes, tokens along failed reasoning paths (red leaves) reduce the value of their prefixes and branching points receive \emph{discriminative signals}, as the value function captures how sibling continuations differ in correctness. By computing $V(s_t)$ directly from the tree, TEMPO provides token-level evaluative feedback while remaining model-free, combining the efficiency of GRPO with the structured credit assignment of PPO.

\subsection{Branch-Aware Advantage Estimation}

Having defined the prefix tree value function $V(s_t)$, we now describe how TEMPO constructs advantages by combining \emph{response-level Monte Carlo signals} from GRPO with \emph{token-level temporal-difference corrections} derived from the tree.

\paragraph{Response-level (MC) signal.}
GRPO provides outcome-level supervision by normalizing the rewards across a group of $G$ responses. For outcome supervision, each response $o_i$ receives a normalized reward
\[
\tilde r_i = \frac{r_i - \text{mean}(r)}{\text{std}(r)},
\]
and assigns it uniformly to all tokens of the trajectory. This yields a pure Monte Carlo signal: every token in a response inherits the same scalar advantage $\tilde r_i$. While efficient, this discards the structure of reasoning trajectories.
\begin{figure*}
    \centering
        % \vspace{-5mm}
    \includegraphics[width=0.8\linewidth]{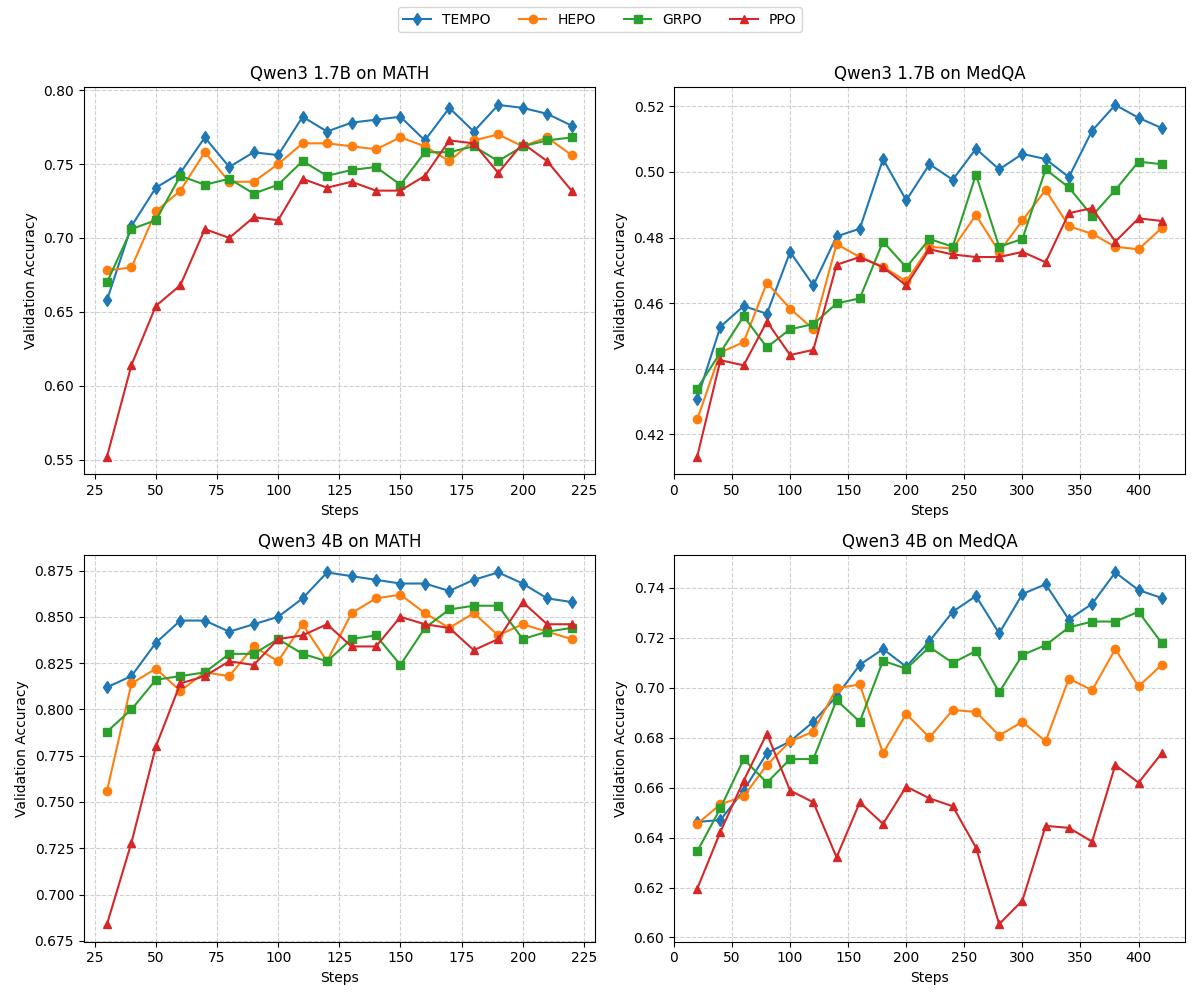}
        % \vspace{-2mm}
  \caption{
    Validation accuracy of MATH and MedQA for Qwen3-1.7B and Qwen3-4B. 
    We compare TEMPO with PPO, GRPO, and HEPO. TEMPO consistently achieves higher accuracy and faster convergence across both domains and model sizes.}
        % \vspace{-5mm}
    \label{fig:val}
\end{figure*}
\paragraph{Token-level (TD) correction.}
TEMPO augments this outcome-level signal with a token-level TD term based on branch-aware values.  
For token $t$ in trajectory $i$, with state prefix $s_t$ and successor $s_{t+1}$, we define the TD error as
\[
\delta_{i,t} = V(s_{t+1}) - V(s_{t}).
\]
This term captures how much the estimated value changes when extending from prefix $s_t$ to $s_{t+1}$. Importantly, $\delta_{i,t}$ is only nonzero at branching points, since non-branch tokens have identical descendant outcomes and thus $V(s_{t+1})=V(s_t)$.

\paragraph{Combined TEMPO advantage.}
The final TEMPO advantage integrates both levels of signal:
\[
\hat A_{i,t} =  \frac{1}{\text{std}(r)} [\underbrace{r_i - \text{mean}(r)}_{\text{MC signal}} \;+\; \underbrace{V(s_{t+1}) - V(s_{t})}_{\text{TD error}}]\]

\begin{itemize}
    \item The \emph{MC component} provides global outcome-level supervision, aligning each response relative to its group.
    \item The \emph{TD error} provides local, branch-aware token-level feedback, highlighting where reasoning paths diverge in quality.
\end{itemize}

% \paragraph{Interpretation.}
% \begin{itemize}
%     \item At non-branch tokens, $\delta_{i,t}=0$ and TEMPO reduces to GRPO.
%     \item At branch tokens, $\delta_{i,t}$ propagates differences in descendant correctness back to the branching point, allowing finer credit assignment.
%     \item Unlike PPO, TEMPO does not train a separate value model: $V(s_t)$ is estimated directly from the group of responses.
% \end{itemize}

\subsection{Policy Update}
For the policy update, TEMPO follows the same principles as DAPO \cite{yu2025dapo}, incorporating several practical design choices that improve stability and efficiency such as Clip-Higher (decoupled clipping), token-level policy-gradient loss (global token averaging) and remove KL divergence.
% \end{itemize}
\paragraph{TEMPO loss function.}
Combining these practices with our proposed branch-aware advantage estimation, the loss is defined as
\begin{align}
\mathcal{J}_{\text{TEMPO}}(\theta)
&=
\mathbb{E}_{\,q,\{o_i\}_{i=1}^G \sim \pi_{\theta_{\mathrm{old}}}(\cdot\mid q)}
\Bigg[
\frac{1}{\sum_{i=1}^{G}|o_i|}
\sum_{i=1}^{G}\sum_{t=1}^{|o_i|}
\min\!\Big(
r_{i,t}(\theta)\,\hat{A}_{i,t}, \nonumber \\
&\hspace{8em}
\mathrm{clip}\!\big(r_{i,t}(\theta),\,1-\varepsilon_{\mathrm{low}},\,1+\varepsilon_{\mathrm{high}}\big)\,\hat{A}_{i,t}
\Big)
\Bigg],
\label{eq:grpo-dapo}
\end{align}

% \begin{table}[t]
% \centering
% \caption{Comparison of PPO, GRPO, and TEMPO.}
% \label{tab:method_compare}
% \begin{tabular}{lcccc}
% \toprule
% \textbf{Method} & \textbf{MC} & \textbf{TD} & \textbf{Token-level} & \textbf{Value Model} \\
% \midrule
% PPO   & Yes & Yes & Yes & Yes \\
% GRPO  & Yes & No  & No  & No  \\
% TEMPO & Yes & Yes & Yes & No  \\
% \bottomrule
% \end{tabular}
% \end{table}

\section{Experimental Setup}

\paragraph{Datasets and Models}
We consider two domains: mathematics and medicine. 
For training, we adopt one representative dataset from each domain: 
MATH~\citep{hendrycks2measuring} for mathematical reasoning and 
MedQA~\citep{jin2021disease} for medical question answering. 
These serve as the \emph{in-distribution (ID)} training tasks. 
For evaluation, we test on both the in-distribution test sets of MATH and MedQA, as well as multiple \emph{out-of-distribution (OOD)} benchmarks to assess generalization. 
In the math domain, OOD benchmarks include GSM-HARD~\citep{gao2023pal}, a challenging variant of GSM8K with harder grade-school problems, and AMC23\footnote{https://huggingface.co/datasets/AI-MO/aimo-validation-amc}, a set of recent American Mathematics Competition problems. 
In the medical domain, OOD benchmarks include MedMCQA~\citep{pal2022medmcqa}, a dataset con-
sisting of multiple-choice medical questions
designed to test clinical knowledge, and MMLU-Medical~\citep{singhal2023large}, a
medical subset of the Massive Multitask Language Understanding (MMLU) benchmark focusing on diverse topics in the medical field. We adopt two publicly available models from the Qwen 3 ~\citep{yang2025qwen3} family: 
Qwen3-1.7B and Qwen3-4B. Both models are fine-tuned in our experiments under identical settings to ensure fair comparison. 

\begin{table*}[t]
\centering
\resizebox{\linewidth}{!}{
\begin{tabular}{l|ccc|ccc}
\toprule
\textbf{Model} & \multicolumn{3}{c|}{\textbf{Math}} & \multicolumn{3}{c}{\textbf{Medical}} \\
\cmidrule(r){2-4} \cmidrule(l){5-7}
& \textbf{MATH} & \textbf{GSM-HARD} & \textbf{AMC23} 
& \textbf{MedQA} & \textbf{MedMCQA} & \textbf{MMLU-Medical} \\
\midrule
Qwen3-1.7B              & 68.5 & 46.85 & 57.5 & 46.11 & 43.17 & 57.85 \\
\quad + PPO             & 81.6 & 53.37 & 67.5 & 52.94 & 48.05 & 70.16 \\
\quad + GRPO            & 82.4 & 53.15 & 72.5 & 56.24 & 49.56 & 71.53 \\
\quad + HEPO            & 81.7 & 52.09 & 62.5 & 54.28 & 48.98 & 71.35 \\
\quad + TEMPO            & \textbf{87.0} & \textbf{56.71} & \textbf{75.0} & \textbf{59.15} & \textbf{51.54} & \textbf{73.37} \\
\midrule
Qwen3-4B                & 71.3 & 54.13 & 75.0 & 65.36 & 56.63 & 78.24 \\
\quad + PPO             & 87.4 & 58.07 & 85.0 & 72.03 & 59.29 & 83.10 \\
\quad + GRPO            & 87.6 & 59.81 & 85.0 & 76.12 & 60.55 & 83.19 \\
\quad + HEPO            & 88.2 & 59.51 & 82.5 & 74.31 & 59.48 & 82.37 \\
\quad + TEMPO            & \textbf{91.0} & \textbf{62.32} & \textbf{92.5} & \textbf{79.18} & \textbf{62.51} & \textbf{85.49} \\
\bottomrule
\end{tabular}
}
\caption{Comparison of PPO, GRPO, HEPO, and TEMPO on mathematical and medical reasoning benchmarks using Qwen3-1.7B and Qwen3-4B as base models. MATH and MedQA are considered \textit{in-distribution} (ID) tasks, while GSM-HARD, AMC23, MedMCQA, and MMLU-Medical are treated as \textit{out-of-distribution} (OOD) evaluations.}
\label{tab:TEMPO_results}
\end{table*}

\paragraph{Baselines}
Our main baseline is GRPO~\citep{shao2024deepseekmath}, which incorporates several practical strategies from DAPO~\citep{yu2025dapo}: removing the KL penalty, introducing a clip-higher mechanism, and applying a token-level policy gradient loss. These modifications make GRPO one of the state-of-the-art RLVF algorithms without requiring a value network. 
\textbf{TEMPO} builds on GRPO and improves credit assignment by exploiting the tree structure of responses. We also compare against a GRPO variant that targets \emph{high-entropy minority tokens}~\citep{wang2025beyond}, where gradient updates are applied only to high-entropy tokens. For clarity in experiments and figures, we denote this variant as HEPO (High Entropy Policy Optimization). Finally, we include an actor–critic baseline: PPO~\citep{schulman2017proximal}, where the critic model is matched in size to the actor model.

\paragraph{Training Details and Hyperparameters} We adopt a binary task reward R that evaluates final answer correctness
against ground truth, following previous work \cite{huang2024n+, ivison2024unpacking}. To ensure fair comparison, all
methods consume the same number of episodes during training: for each question, we sample 6 episodes and go
over the dataset 10 times, yielding 60 episodes per question
across all methods.

\section{Results}
In this section, we evaluate the effect of better Credit Assignment on task
performance, efficiency, and generalization dynamics.

\subsection{Task Performance}

% Figure~\ref{fig:val} presents validation curves on MATH and MedQA. Across both Qwen3-1.7B and Qwen3-4B, \textbf{TEMPO} outperforms all baselines in terms of both convergence speed and final accuracy. 
% On MATH, TEMPO converges significantly faster than PPO and GRPO, reaching high accuracy within fewer steps, and ultimately surpasses them by $2{-}3$ points.  On MedQA, a more challenging dataset, the gap is even more pronounced: PPO exhibits unstable training and low final accuracy, while GRPO and HEPO improve stability but plateau earlier. TEMPO continues to climb and achieves the best validation performance, demonstrating that tree-structured TD corrections enhance efficiency and generalization in medical reasoning tasks.
% \paragraph{Results.} 
Figure~\ref{fig:val} shows validation curves on MATH and MedQA for both Qwen3-1.7B and Qwen3-4B. Across all settings, \textbf{TEMPO} achieves the best performance in terms of convergence speed and final accuracy. On \textbf{MATH}, TEMPO consistently outperforms other methods, followed by HEPO, then GRPO and PPO. The fact that HEPO performs slightly better than GRPO and PPO suggests that focusing updates on high-entropy tokens helps exploit the reasoning structures already present in the model. 
Since mathematical reasoning knowledge is largely captured during pretraining and instruction tuning, the RL process primarily reinforces existing structures rather than learning new ones. 
Thus, token-structure–oriented methods like HEPO gain an advantage compare with GRPO. On \textbf{MedQA}, however, the trend differs. TEMPO again delivers the best results, but GRPO surpasses HEPO, and PPO lags behind all others. We hypothesize that medical reasoning requires learning novel domain-specific knowledge, which emphasizes \emph{exploration} rather than pure exploitation of existing token-level structures. 
Here, GRPO’s group-relative normalization provides stronger signals than PPO, while HEPO’s focus on high-entropy tokens is insufficient to capture new knowledge. TEMPO combines the benefits of GRPO with tree-structured TD guidance, enabling effective exploration while still leveraging structural signals, leading to the best generalization in the medical domain.

\begin{figure*}
    \centering
    \includegraphics[width=0.9\linewidth]{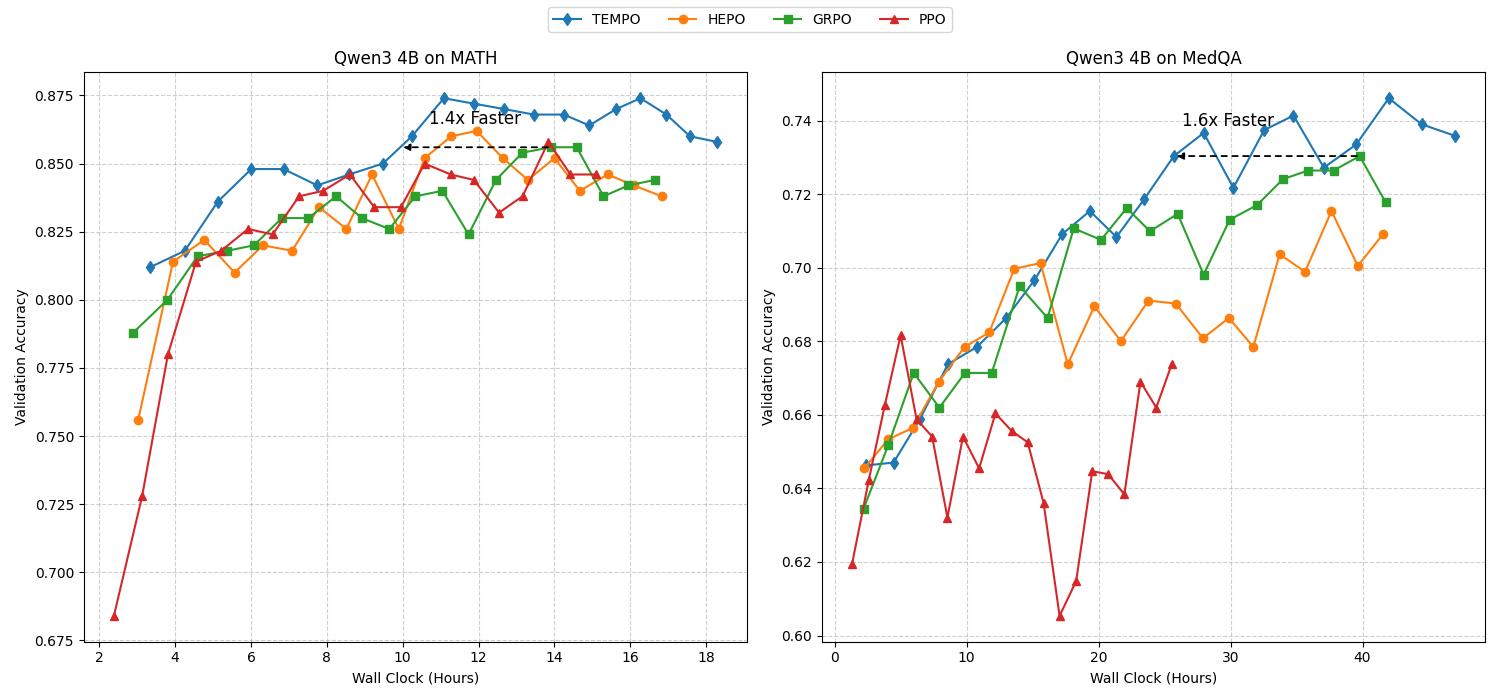}
    \caption{TEMPO converges faster and to higher accuracy than GRPO, passes GRPO’s peak performance in fewer iterations and less overall time.}
    \label{fig:timing}
\end{figure*}

\subsection{Computational Efficiency}
Figure~\ref{fig:timing} reports validation accuracy against wall-clock training time on Qwen3-4B for both MATH and MedQA, with all runs executed on identical hardware (2$\times$NVIDIA H100 80GB GPUs). 
PPO shows the slowest convergence and lowest final accuracy, while GRPO and HEPO provide more stable training but plateau earlier. 
In contrast, \textbf{TEMPO demonstrates clear computational advantages}: on MATH, TEMPO achieves GRPO’s best accuracy about \textbf{1.4$\times$ faster}, and on MedQA it reaches GRPO’s peak roughly \textbf{1.6$\times$ faster}. 
Moreover, TEMPO continues to improve beyond these points, ultimately attaining higher final accuracy. 
These results show that integrating tree-structured TD corrections improves both credit assignment and training efficiency under realistic hardware budgets.

\subsection{Effect of Group Size}

\begin{wrapfigure}{r}{0.4\textwidth}
    \centering
    \vspace{-1.2cm}
    \includegraphics[width=0.95\linewidth]{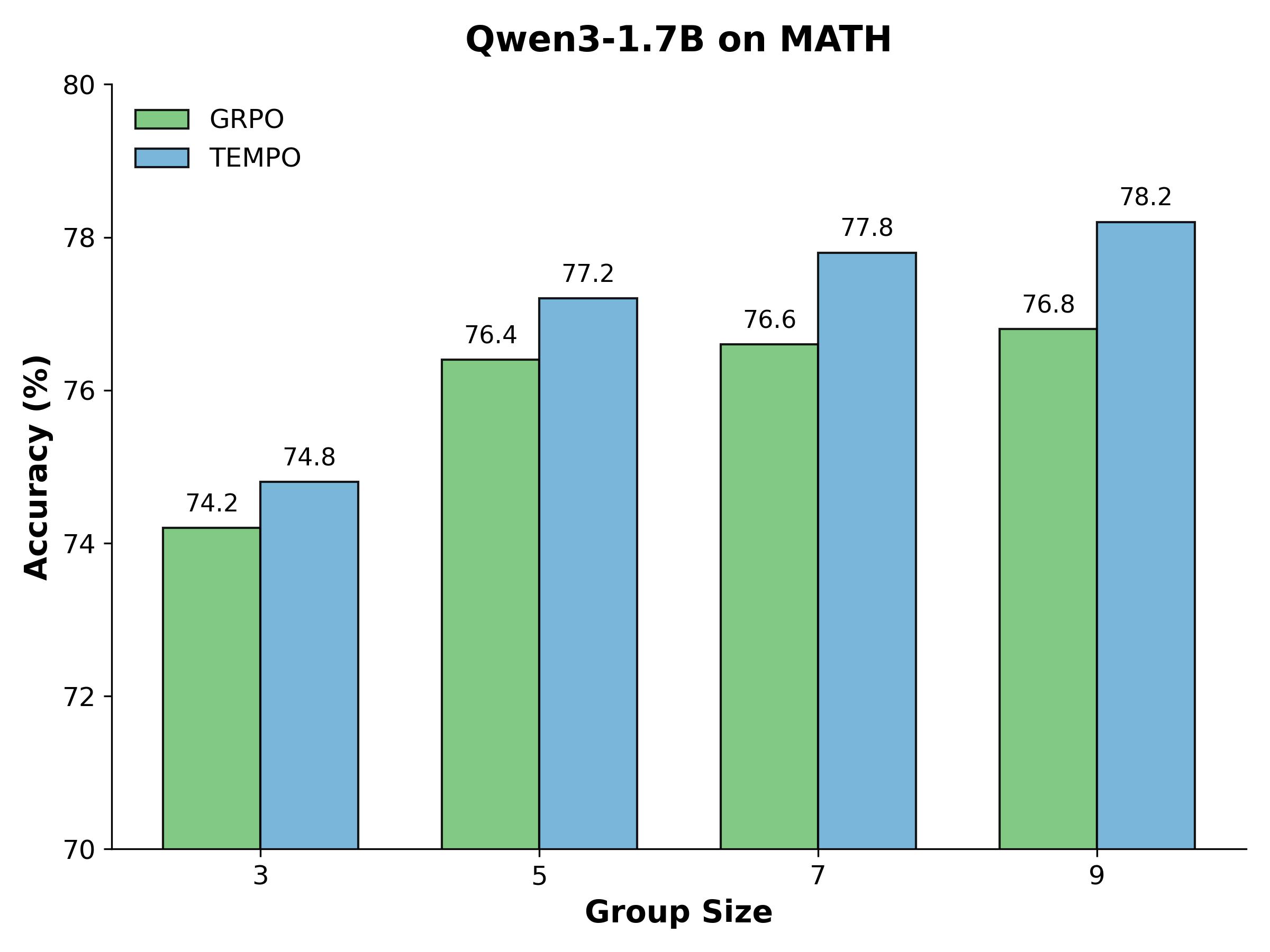}
    \caption{Effect of group size on MATH accuracy for Qwen3-1.7B. TEMPO consistently outperforms GRPO across all settings.}
    \label{fig:groupsize}
    % \vspace{-0.1cm}
\end{wrapfigure}

An important hyperparameter in GRPO-style methods is the \emph{group size}, i.e., the number of responses sampled per prompt during training. 
Larger groups provide a more reliable relative baseline, but also increase computational cost. Figure~\ref{fig:groupsize} shows the effect of varying group size ($3, 5, 7, 9$) on MATH accuracy for Qwen3-1.7B. We observe two key trends. First, increasing group size improves performance for both GRPO and TEMPO, consistent with prior findings that larger groups yield stronger learning signals. Second, \textbf{TEMPO consistently outperforms GRPO across all group sizes}, with gains of around $0.5{-}2$ points in accuracy. 
This demonstrates that tree-structured TD corrections complement group-relative normalization and remain effective even with small groups, making TEMPO more robust under limited sampling budgets. 

\begin{wrapfigure}[14]{r}{0.5\textwidth}
  \centering
  \vspace{-1.2cm}
  \includegraphics[width=\linewidth]{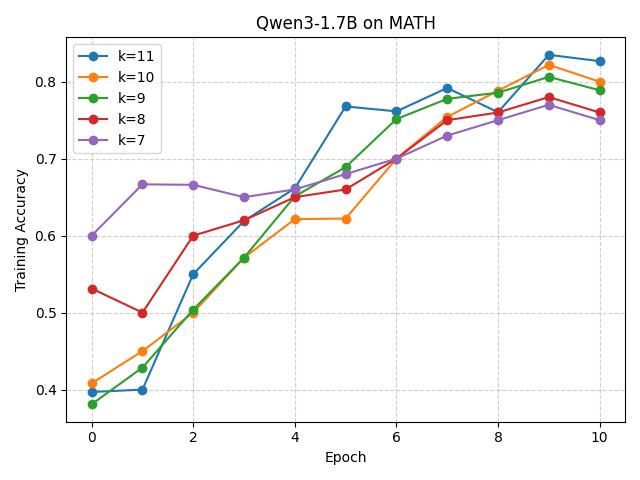}
  \caption{Training accuracy vs.\ epochs for different numbers of preserved branches $k$ at group size $G{=}7$.}
  \label{fig:branch-k}
  % \vspace{-0.5cm}
\end{wrapfigure}

\subsection{Effect of Branch Count}
To study the effect of branching toward training performance, we build the prefix tree at the first epoch and record the number of branches $k$.  
Responses prefix trees are then grouped by their initial $k\!\in\!\{7,8,9,10,11\}$, and we track their training accuracy over subsequent epochs.  
As shown in Figure~\ref{fig:branch-k}, responses that begin with more branches learn faster and reach higher accuracy, reflecting the availability of more branch tokens where TEMPO applies non-zero TD corrections. When the observed branching is minimal ($k{=}G{=}7$), the TD signal largely vanishes and TEMPO behaves like GRPO, yielding the slowest improvement among the others.

\begin{figure}[t]
  \centering
  \begin{subfigure}{0.49\linewidth}
    \centering
    \includegraphics[width=\linewidth]{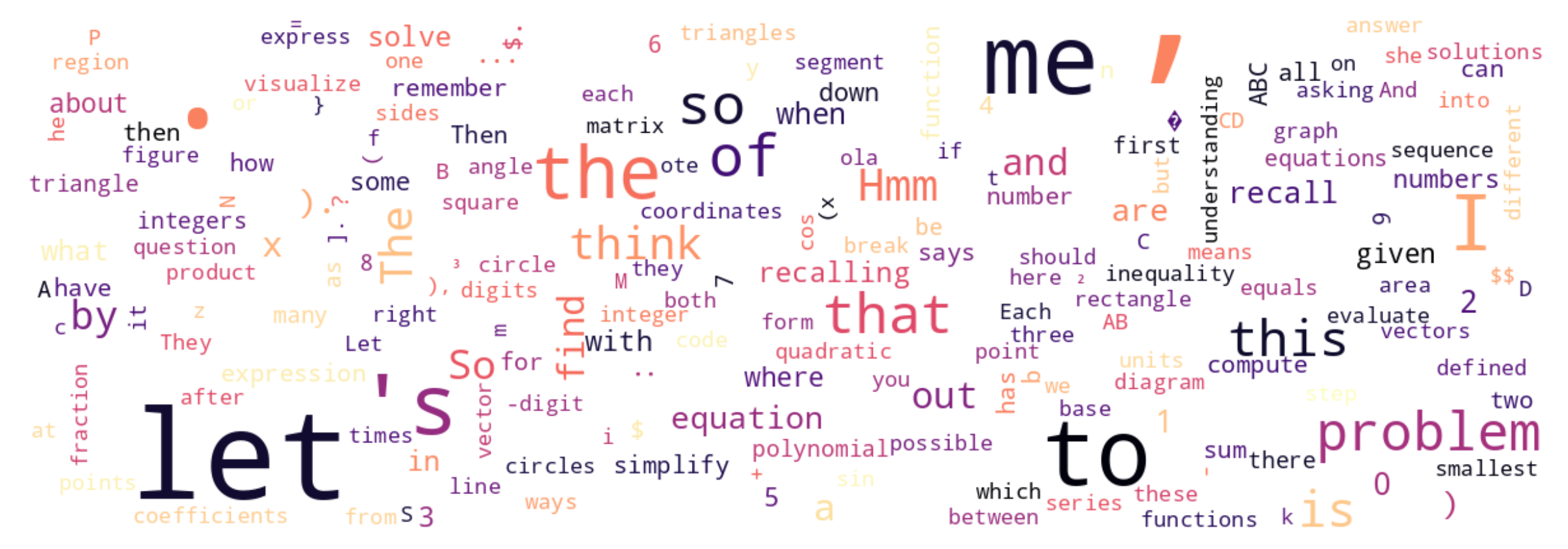}
    \caption{MATH (branching tokens)}
    \label{fig:wc-math}
  \end{subfigure}\hfill
  \begin{subfigure}{0.49\linewidth}
    \centering
    \includegraphics[width=\linewidth]{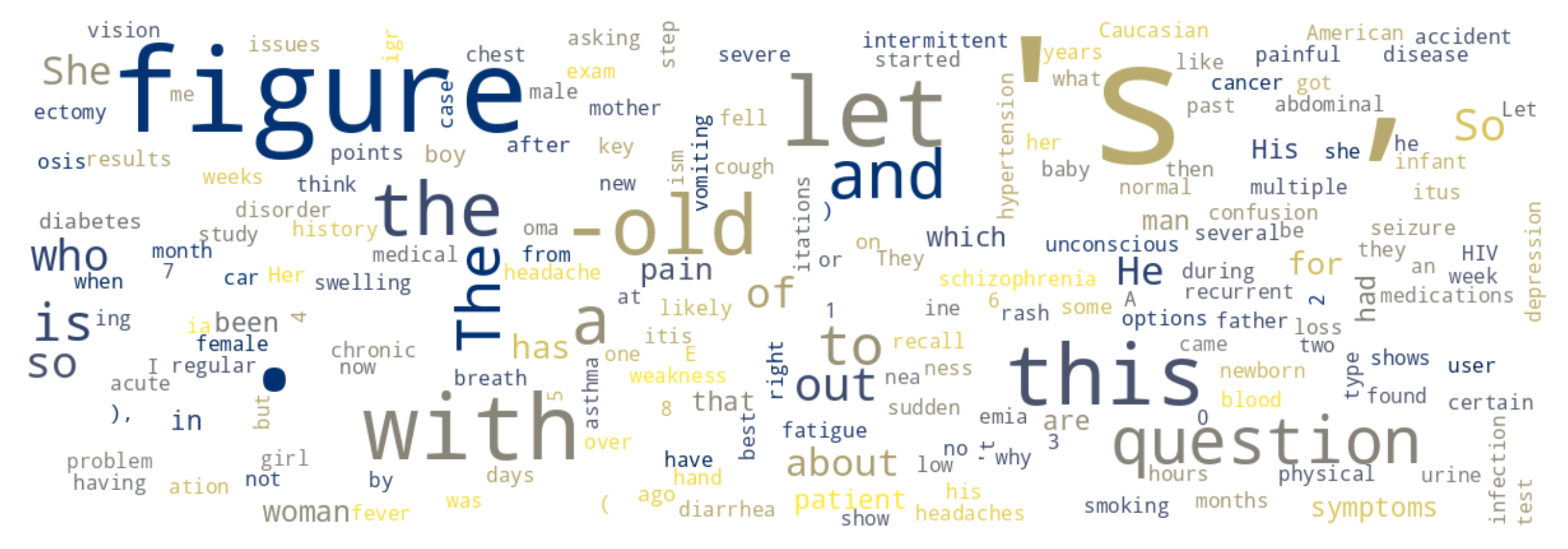}
    \caption{MedQA (branching tokens)}
    \label{fig:wc-med}
  \end{subfigure}
  \caption{Word clouds of \emph{branching tokens} (high-entropy / multi-child nodes) collected during TEMPO training. Token sizes reflect frequency among branch points across training steps. In both domains, branching tokens align with decision points where the reasoning path can fork.}
  \label{fig:wc-both}
\end{figure}

\subsection{Where Do Branches Happen?}

Figure~\ref{fig:wc-both} visualizes the tokens that most often appear at \emph{branch nodes}—prefixes $s_t$ with multiple continuations or high next-token entropy—aggregated over TEMPO training. On Math, branching tokens cluster around \emph{planning and formalization} phrases such as ``let'', ``find'', ``solve'', ``suppose'', numerals, variables, and operators (e.g., ``$x$'', ``equation'', ``series'', ``triangle''). These are the points where the model chooses a solution strategy (set up variables, recall a fact, pick a formula) before committing to derivations. TEMPO’s TD correction therefore acts exactly where the plan can diverge (e.g., choosing the wrong identity vs.\ the right one). In contrast, medical branching tokens emphasize \emph{clinical entities and constraints}: demographics (``year-old'', ``man/woman''), symptoms (``pain'', ``fever'', ``cough''), disease terms (``diabetes'', ``seizure''), and linking words that steer differentials (``with'', ``who'', ``which''). These tokens define the candidate diagnosis/workup branches, so TEMPO focuses signal where the case interpretation can split. Across domains, branches coincide with \textbf{high-stakes decision tokens}—the junctures that determine the downstream trajectory. This qualitative evidence complements our quantitative results: the tree-aware TD signal is delivered exactly where it matters most.

% \paragraph{Takeaways.}
% Across domains, branches coincide with \textbf{high-stakes decision tokens}—the junctures that determine the downstream trajectory. By concentrating TD corrections on these tokens (while leaving non-branch tokens GRPO-like), TEMPO supplies precise, structure-aware credit that improves both convergence speed and generalization. This qualitative evidence complements our quantitative results: the tree-aware TD signal is delivered exactly where it matters most.

% requires \usepackage{wrapfig}

\subsection{Generalization}

Table~\ref{tab:TEMPO_results} summarizes performance on both in-distribution (ID) and out-of-distribution (OOD) benchmarks for mathematics and medicine. 
On the ID tasks (MATH and MedQA), \textbf{TEMPO consistently achieves the highest accuracy}, surpassing PPO, GRPO, and HEPO across both Qwen3-1.7B and Qwen3-4B. For example, TEMPO improves MedQA accuracy from 76.1\% (GRPO, 4B) to 79.2\%, and raises MATH accuracy from 87.6\% to 91.0\%.  On OOD evaluations, TEMPO also establishes clear gains. In the math domain, it pushes GSM-HARD accuracy from 59.8\% (GRPO, 4B) to 62.3\%, and AMC23 from 85.0\% to 92.5\%. In the medical domain, it improves MedMCQA from 60.55\% to 62.51\% and MMLU-Medical from 83.2\% to 85.5\%. These improvements across unseen distributions highlight that TEMPO not only enhances in-distribution learning efficiency but also yields stronger generalization to harder and more diverse reasoning tasks. 

\begin{wrapfigure}{r}{0.55\textwidth}
    \centering
    \vspace{-0.5cm}
    \includegraphics[width=0.5\textwidth]{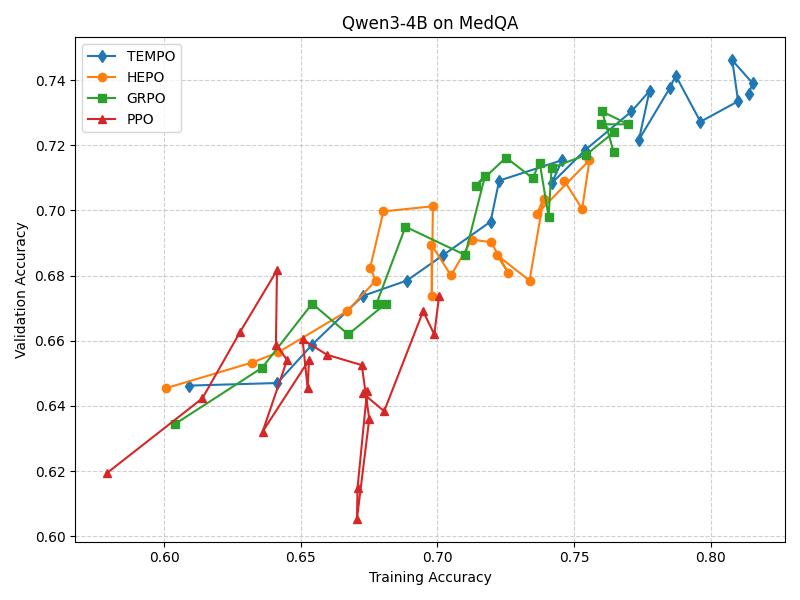}
    \caption{Training vs. validation accuracy on MedQA with Qwen3-4B. PPO overfits to training data, while TEMPO maintains better generalization.}
    \label{fig:overfit}
    \vspace{-0.6cm}
\end{wrapfigure}

To further investigate the gap, we analyze the relationship between training and validation accuracy on MedQA for Qwen3-4B (Figure~\ref{fig:overfit}). 
We find that PPO exhibits clear overfitting: its training accuracy continues to increase while validation accuracy plateaus, indicating weak generalization. In contrast, both GRPO and TEMPO improve training and validation accuracy in tandem, with TEMPO achieving the highest performance on both, suggesting more reliable generalization rather than memorization of the training distribution. This analysis explains why PPO lags behind in final MedQA accuracy and underscores TEMPO’s advantage when scaling to larger models and more challenging domains.

\section{Conclusion}

In this work, we introduced \textbf{TEMPO}, a reinforcement learning algorithm for LLM alignment that integrates temporal-difference (TD) signals into group-relative optimization by exploiting the tree structure of sampled responses. Unlike PPO, which requires training a separate value network, and GRPO, which discards token-level information by relying purely on Monte Carlo signals, TEMPO unifies the strengths of both approaches without additional model components. By deriving value estimates directly from the response tree, TEMPO enables token-level TD corrections on top of group-relative normalization, yielding more fine-grained and stable credit assignment. Our experiments across mathematics and medicine demonstrate two key findings. 
First, TEMPO achieves higher accuracy than PPO, GRPO, and HEPO on both in-distribution and out-of-distribution benchmarks, showing strong generalization. Second, TEMPO converges significantly faster in wall-clock time, achieving comparable or better accuracy up to $1.6\times$ earlier under the same hardware configuration. Overall, TEMPO establishes a practical and scalable approach to reinforcement learning with verifiable feedback. 
It provides fine-grained credit assignment without the overhead of a value model, improves training efficiency, and enhances robustness across domains. 
We believe TEMPO opens up new directions for reasoning-focused LLM training, and future work may extend tree-structured TD signals to multi-step verification, retrieval-augmented reasoning, or other structured alignment settings.

\section*{Acknowledgments}

This material is the result of work supported with resources and the use of facilities at the Center for Healthcare Organization and Implementation Research, VA Bedford Health Care.

\bibliography{iclr2026_conference}
\bibliographystyle{iclr2026_conference}

\appendix
\section{Appendix}

\subsection{Implementation Details}
To ensure our GRPO implementation is robust, and our evaluation reflects its full potential, we have applied a set of wellestablished techniques and best practices from the literature \cite{yu2025dapo}. Below, we outline the key implementation details that were most effective in our experiments:
\begin{itemize}
    \item Clip-Higher (decoupled clipping). We decouple the clipping bounds and raise the upper cap $(1+\varepsilon_{\text{high}})$ while keeping the lower cap $(1-\varepsilon_{\text{low}})$, which allows low-probability “exploration” tokens to increase more freely and helps prevent entropy collapse.
    \item Token-level policy-gradient loss: Token-level policy-gradient loss (global token averaging): We optimize a token-level surrogate averaged over \emph{all} tokens in the batch, broadcasting each response’s group-normalized outcome reward to its tokens since sample-level averaging underweights long responses and fails to penalize low-quality long patterns, which destabilizes training; token-level loss restores balanced credit assignment and yields healthier length/entropy dynamics.
    \item  Remove KL divergence: In long-CoT reasoning, the online policy can beneficially diverge from the initialization; thus we omit an explicit KL regularizer and rely on clipping for stability.
\end{itemize}

\subsection{Hyperparameters}

In this section, we provide a comprehensive overview of the hyperparameters used in our experiments. The number of training episodes was carefully selected to ensure that the amount of training data remained consistent across all methods.

\paragraph{PPO}
Finetuning LLMs with PPO is known to be highly sensitive to hyperparameter choices, making optimal selection critical for strong performance. To ensure robustness, we considered hyperparameter values reported in prior studies~\cite{shao2024deepseekmath} and performed extensive sweeps across a wide range of candidate values. Specifically, we first identified the set of hyperparameters that achieved the best performance across both the MATH and MedQA tasks using the Qwen3 1.7B model. This optimal configuration was then employed for the remainder of our experiments. The complete list of PPO hyperparameters, along with their respective search spaces, is shown in Table~\ref{tab:ppo_hparams}.

\paragraph{GRPO, HEPO, and TEMPO}
Since policy optimization in RLOO and GRPO closely resembles PPO, we initialized their hyperparameters using the PPO configuration. This ensures a strong starting point while enabling a systematic comparison among the algorithms.  We note that the absence of explicit credit assignment in these methods may result in high-variance policy gradient updates, potentially leading to instability~\cite{greensmith2004variance}. The full list of hyperparameters for GRPO, HEPO, and TEMPO is provided in Table~\ref{tab:ppo_hparams}.

\subsection{Compute}
All experiments were conducted using multi-GPU training to efficiently handle the computational demands of large-scale models. For the Qwen3-1.7B model, we utilized a node with 1 × Nvidia H100 80GB GPUs to train both TEMPO and all the baselines. For the larger Qwen3-4B model, we employed a more powerful setup, using a node with 2 × Nvidia H100 80GB GPUs.

\subsection{Software Stack}
For model implementation, we utilize the Huggingface
library. Training is carried out using the VERL \cite{zhang2024framework} distributed training library, which offers efficient multi-GPU support. For trajectory sampling during RL training, we rely on the vLLM library \cite{kwon2023efficient}, which provides optimized inference for LLMs.

\subsection{Reproducity}
In this study, all experiments were conducted using open-source libraries, publicly available datasets, and open-weight LLMs. To ensure full reproducibility, we will make our codebase publicly available on GitHub at \url{https://github.com/fatebreaker/tempo}.

\begin{table*}[t]
\centering

\setlength{\tabcolsep}{10pt}
\renewcommand{\arraystretch}{1.15}
\begin{tabular}{l l l}
\toprule
\textbf{Parameter} & \textbf{Value} & \textbf{Notes} \\
\midrule
\multicolumn{3}{c}{\textbf{Training}} \\
\midrule
Optimizer & AdamW & \\
Adam parameters $(\beta_1,\beta_2)$ & $(0.9,\;0.999)$ & \\
Learning rate & $1 \times 10^{-6}$ & \\
Weight decay & $0.0$ & \\
Warmup & $0\%$ of training steps & \\
\# Train steps (MATH) & 220 steps & $\sim$10 dataset epochs \\
\# Train steps (MedQA) & 420 steps & $\sim$10 dataset epochs \\
\midrule
\multicolumn{3}{c}{\textbf{General}} \\
\midrule
Maximum response length & 1024 tokens & \\
Max seq length & 2048 tokens & \\
\midrule
\multicolumn{3}{c}{\textbf{PPO}} \\
\midrule
Mini-batch size & 64 & \\
\# Inner epochs per PPO step & 2  \\
Discount factor $\gamma$ & 1.0 & \\
GAE parameter $\lambda$ & 1.0  \\
KL penalty coefficient $\beta$ & $1\!\times\!10^{-4}$  \\
\multicolumn{3}{c}{\textbf{GRPO/HEPO/TEMPO}} \\
\midrule
\# Responses per prompt & 6 \\
Mini-batch size & 64 & \\
Discount factor $\gamma$ & 1.0 & \\
KL penalty coefficient $\beta$ & 0.0 \\
Policy clipping parameter $\epsilon$ & 0.28, 0.2 & \\
\multicolumn{3}{c}{\textbf{HEPO}} \\
\midrule
$\rho$ & 0.2 & Only do gradient update on top 20\% high entroy tokens \\
\bottomrule
\end{tabular}
\caption{Summary of hyperparameters used in the experiments.}
\label{tab:ppo_hparams}
\end{table*}

% \end{itemize}

\end{document}